\documentclass[runningheads, envcountsame, a4paper]{llncs}
\usepackage[pdftex]{graphicx}
\usepackage{epstopdf}
\usepackage[hyphens]{url}
\usepackage[misc]{ifsym}
\usepackage{microtype}
\usepackage{booktabs}       % professional-quality tables

\usepackage{amsfonts}       % blackboard math symbols
\usepackage{nicefrac}       % compact symbols for 1/2, etc.
\usepackage{microtype}      % microtypography
\usepackage{amsmath}
\usepackage[boldmath]{numprint}
\usepackage[mathscr]{eucal}
\usepackage{algorithm}
\usepackage{algorithmic}
\usepackage{wrapfig}
\usepackage{filecontents}
\usepackage[caption=false]{subfig} 
\begin{document}
\title{Generating Black-Box Adversarial Examples for Text Classifiers Using a Deep Reinforced Model}
\titlerunning{Black-Box Adversarial Examples for Text Classifiers}
\toctitle{Generating Black-Box Adversarial Examples for Text Classifiers Using a Deep Reinforced Model}

\author{Prashanth Vijayaraghavan \Letter \and
Deb Roy}
\authorrunning{P. Vijayaraghavan et al.}
% \institute{
\tocauthor{Prashanth~Vijayaraghavan, Deb~Roy}

\institute{MIT Media Lab, Cambridge MA 02139, USA   \\
% \url{https://www.media.mit.edu/} \\
\email{$\{$pralav,dkroy$\}$@media.mit.edu}
}

\maketitle              % typeset the header of the contribution
\begin{abstract}

Recently, generating adversarial examples has become an important means of measuring robustness of a deep learning model. Adversarial examples help us identify the susceptibilities of the model and further counter those vulnerabilities by applying adversarial training techniques. In natural language domain, small perturbations in the form of misspellings or paraphrases can drastically change the semantics of the text. We propose a reinforcement learning based approach towards generating adversarial examples in black-box settings. We demonstrate that our method is able to fool well-trained models for (a) IMDB sentiment classification task and (b) AG's news corpus news categorization task with significantly high success rates. We find that the adversarial examples generated are semantics-preserving perturbations to the original text. 
\end{abstract}

\section{Introduction}
\label{intro}
Adversarial examples are generally minimal perturbations applied to the input data in an effort to expose the regions of the input space where a trained model performs poorly.
Prior works \cite{biggio2013evasion,szegedy2013intriguing} have demonstrated the ability of an adversary to evade state-of-the-art classifiers by carefully crafting attack examples which can be even imperceptible to humans. Following such approaches, there has been a number of techniques aimed at generating adversarial examples \cite{moosavi2016deepfool,xiao2018spatially}. Depending on the degree of access to the target model, an adversary may operate in one of the two different settings: (a) black-box setting, where an adversary doesn't have access to target model's internal architecture or its parameters, (b) white-box setting, where an adversary has access to the target model, its parameters, and input feature representations. In both these settings, the adversary cannot alter the training data or the target model itself. 
 Depending on the purpose of the adversary, adversarial attacks can be categorized as (a) targeted attack and (b) non-targeted attack.
In a targeted attack, the output category of a generated example is intentionally controlled to a specific target category with limited change in semantic information. While a non-targeted attack doesn't care about the category of misclassified results.

Most of the prior work has focused on image classification models where adversarial examples are obtained by introducing imperceptible changes to pixel values through optimization techniques \cite{kurakin2016adversarial,iter2017generating}. However, generating natural language adversarial examples can be challenging mainly due to the discrete nature of text samples. Continuous data like image or speech is much more tolerant to perturbations compared to text \cite{goodfellow2015explaining}. In textual domain, even a small perturbation is clearly perceptible and can completely change the semantics of the text. Another challenge for generating adversarial examples relates to identifying salient areas of the text where a perturbation can be applied successfully to fool the target classifier. In addition to fooling the target classifier, the adversary is designed with different constraints depending on the task and its motivations \cite{gilmer2018motivating}. In our work, we focus on constraining our adversary to craft examples with semantic preservation and minimum perturbations to the input text. 
 
 % In a black-box setting, several studies have relied on heuristic approaches for defining adversarial perturbations on text classifiers. The steps involved in crafting adversarial sequences include: (a) choosing tokens/regions of text either randomly or using scoring functions to perturb, (b) introducing perturbations to the chosen token in the form of like misspellings, paraphrases or modifications in the embedding space of the word. 
Given different settings of the adversary, there are other works that have designed attacks in ``gray-box'' settings \cite{biggio2018wild,guo2017countering,mopuri2018gray}. However, the definitions of ``gray-box'' attacks are quite different in each of these approaches. In this paper, we focus on ``black-box'' setting where we assume that the adversary possesses a limited set of labeled data, which is different from the target's training data, and also has an oracle access to the system, i.e., one can query the target classifier with any input and get its corresponding predictions. We propose an effective technique to generate adversarial examples in a black-box setting. We develop an Adversarial Example Generator (AEG) model that uses a reinforcement learning framing to generate adversarial examples. We evaluate our models using a word-based \cite{kim2014convolutional} and character-based \cite{zhang2015character} text classification model on benchmark classification tasks: sentiment classification and news categorization. The adversarial sequences generated are able to effectively fool the classifiers without changing the semantics of the text. Our contributions are as follows:

 \begin{itemize}
 \item We propose a black-box non-targeted attack strategy by combining ideas of substitute network and adversarial example generation. We formulate it as a reinforcement learning task.
 
\item We introduce an encoder-decoder that operates over words and characters of an input text and empowers the model to introduce word and character-level perturbations.

\item We adopt a self-critical sequence training technique to train our model to generate examples that can fool or increase the probability of misclassification in text classifiers.

\item We evaluate our models on two different datasets associated with two different tasks: IMDB sentiment classification and AG's news categorization task. We run ablation studies on various components of the model and provide insights into decisions of our model.
\end{itemize}

\section{Related Work}
Generating adversarial examples to bypass deep learning classification models have been widely studied. In a white-box setting, some of the approaches include gradient-based \cite{kereliuk2015deep,goodfellow2015explaining}, decision function-based \cite{moosavi2016deepfool} and spatial transformation based perturbation techniques\cite{xiao2018spatially}. In a black-box setting, several attack strategies have been proposed based on the property of transferability \cite{szegedy2013intriguing}. Papernot et al. \cite{papernot2016practical,papernot2016transferability} relied on this transferability property where adversarial examples, generated on one classifier, are likely to cause another classifier to make the same mistake, irrespective of their architecture and training dataset. In order to generate adversarial samples, a local substitute model was trained with queries to the target model. Many learning systems allow query accesses to the model. However, there
is little work that can leverage query-based access to target models to construct adversarial samples and move beyond transferability. These studies have primarily focused on image-based classifiers and cannot be directly applied to text-based classifiers.
% In contrast, we show that the proposed AdvGAN can perform black-box attacks without depending on transferability. 

 While there is limited literature for such approaches in NLP systems, there have been some studies that have exposed the vulnerabilities of neural networks in text-based tasks like machine translations and question answering. Belinkov and Bisk  \cite{belinkov2017synthetic} investigated the sensitivity of neural machine translation (NMT) to synthetic and natural noise containing common misspellings. They demonstrate that state-of-the-art models are vulnerable to adversarial attacks even after a spell-checker is deployed. Jia et al. \cite{jia2017adversarial} showed that networks trained for more difficult tasks, such as question answering, can be easily fooled by introducing distracting sentences into text, but these results do not transfer obviously to simpler text classification tasks. Following such works, different methods with the primary purpose of crafting adversarial example have been explored. Recently, a work by Ebrahimi et al. \cite{ebrahimi2017hotflip} developed a gradient-based optimization method that manipulates discrete text structure at its one-hot representation to generate adversarial examples in a white-box setting. In another white-box based attack, Gong et al. \cite{gong2018adversarial} perturbed the word embedding of given text examples and projected them to the nearest neighbour in the embedding space. This approach is an adaptation of perturbation algorithms for images. Though the size and quality of embedding play a critical role, this targeted attack technique ensured that the generated text sequence is intelligible. 
 
 Alzantot et al. \cite{alzantot2018generating} proposed a black-box targeted attack using a population-based optimization via genetic algorithm \cite{anderson1994genetic}. The perturbation procedure consists of random selection of words, finding their nearest neighbours, ranking and substitution to maximize the probability of target category. In this method, random word selection in the sequence to substitute were full of uncertainties and  might be meaningless for the target label when changed. Since our model focuses on black-box non-targeted attack using an encoder-decoder approach, our work is closely related to the following techniques in the literature: Wong (2017) \cite{wong2017dancin}, Iyyer et al. \cite{iyyer2018adversarial} and  Gao et al. \cite{gao2018black}. Wong (2017) \cite{wong2017dancin} proposed a GAN-inspired method to generate adversarial text examples targeting black-box classifiers. However, this approach was restricted to binary text classifiers. Iyyer et al. \cite{iyyer2018adversarial} crafted adversarial examples using their proposed
 Syntactically Controlled Paraphrase Networks (SCPNs). They designed this model for generating syntactically adversarial examples without compromising on the quality of the input semantics. The general process is based on the encoder-decoder architecture of SCPN. Gao et al. \cite{gao2018black} implemented an algorithm called DeepWordBug that generates small text perturbations in a black box setting forcing the deep learning model to make mistakes. DeepWordBug used a scoring function to determine important tokens and then applied character-level transformations to those tokens. Though the algorithm successfully generates adversarial examples by introducing character-level attacks, most of the introduced perturbations are constricted to misspellings. The semantics of the text may be irreversibly changed if excessive misspellings are introduced to fool the target classifier. While SCPNs and DeepWordBug primary rely only on paraphrases and character transformations respectively to fool the classifier, our model uses a hybrid word-character encoder-decoder approach to introduce both paraphrases and character-level perturbations as a part of our attack strategy. Our attacks can be a test of how robust the text classification models are to word and character-level perturbations.

\section{Proposed Attack Strategy}
Let us consider a target model $T$ and $(x,l)$ refers to the samples from the dataset. Given an instance $x$, the goal of the adversary is to generate adversarial examples $x'$ such that $T(x') \neq l$, where $l$ denotes the true label i.e take one of the $K$ classes of the target classification model. The changes made to $x$ to get $x'$ are called perturbations. We would like to have $x'$ close to the original instance $x$. In a black box setting, we do not have knowledge about the internals of the target model or its training data. Previous work by Papernot et al. \cite{papernot2016practical} train a separate substitute classifier such that it can mimic the decision boundaries of the target classifier. The substitute classifier is then used to craft adversarial examples. While these techniques have been applied for image classification models, such methods have not been explored extensively for text. 

We implement both the substitute network training and adversarial example generation using an encoder-decoder architecture called \textit{Adversarial Examples Generator (AEG)}. The encoder extracts the character and word information from the input text and produces hidden representations of words considering its sequence context information. A substitute network is not implemented separately but applied using an attention mechanism to weigh the encoded hidden states based on their relevance to making predictions closer to target model outputs. The attention scores provide certain level of interpretability to the model as the regions of text that need to perturbed can be identified and visualized. The decoder uses the attention scores obtained from the substitute network, combines it with decoder state information to decide if perturbation is required at this state or not and finally emits the text unit (a text unit may refer to a word or character). Inspired by a work by Luong et al. \cite{luong2016achieving}, the decoder is a word and character-level recurrent network employed to generate adversarial examples. Before the substitute network is trained, we pretrain our encoder-decoder model on common misspellings and paraphrase datasets to empower the model to produce character and word perturbations in the form of misspellings or paraphrases. For training substitute network and generation of adversarial examples, we randomly draw
data that is disjoint from the training data of the black-box
model since we assume the adversaries have no prior knowledge about the training data or the model. Specifically, we consider attacking a target classifier by generating adversarial examples based on unseen input examples. This is done by dividing the dataset into training, validation and test  using 60-30-10 ratio. The training data is used by the target model, while the unseen validation samples are used with necessary data augmentation for our \textit{AEG} model. We further improve our model by using a self-critical approach to finally generate better adversarial examples. The rewards are formulated based on the following goals: (a) fool the target classifier, (b) minimize the number of perturbations and (c) preserve the semantics of the text. In the following sections, we explain the encoder-decoder model and then describe the reinforcement learning framing towards generation of adversarial examples.

\label{datasets}
\subsection{Background and Notations}
\label{stdseq}
Most of the sequence generation models follow an encoder-decoder framework \cite{sutskever2014sequence,chung2014empirical,kalchbrenner2013recurrent} where encoder and decoder are modelled by separate recurrent neural networks. Usually these models are trained using a pair of text $(x,y)$ where $x=[x_1, x_2..,x_n]$ is the input text and the $y=[y_1, y_2..,y_m]$ is the target text to be generated. The encoder transforms an input text sequence into an abstract representation $h$. While the decoder is employed to generate the target sequence using the encoded representation $h$. 
% Specifically, the target units (words or characters) are generated sequentially by jointly conditioning on the encoded input representation $h$ and previously generated units and passing it on to a feed-forward layer. We express feed-forward layers as $f_{(\cdot)}(s,h)= A(W_as+U_ah)$, where $A$ is the activation function like $softmax$ or $tanh$ depending on its usage context. The number of terms inside the activation function depends on number of parameters given to $f_{(\cdot)}$.
However, there are several studies that have incorporated several modifications to the standard encoder-decoder framework \cite{bahdanau2014neural,luong2016achieving,luong2014addressing}. 

% The log likelihood of the decoded text, one unit at a time, can be computed using:
% \begin{equation}
%     log p(y|x)=\sum_{j=1}^{m}{log p(y_j|y_{<j},h)}
%     \label{likelihood}
% \end{equation}
\paragraph{Encoder}
Based on Bahdanau et al. \cite{bahdanau2014neural}, we encode the input text sequence using bidirectional gated recurrent units (GRUs) to encode the input text sequence $x$. Formally, we obtain an encoded representation given by: $\overleftrightarrow{h_t}= \overleftarrow{h_t} + \overrightarrow{h_t}$.

\paragraph{Decoder}
\label{decoder}
The decoder is a forward GRU implementing an attention mechanism to recognize the units of input text sequence relevant for the generation of the next target work. The decoder GRU generates the next text unit at time step $j$ by conditioning on the current decoder state $s_j$, context vector $c_j$ computed using attention mechanism and  previously generated text units. The probability of decoding each target unit is given by:

\begin{gather}
p(y_j|y_{<j},h)=softmax(\tilde{s}_j)\\
\tilde{s}_j=f_d([c_j; s_j]) 
\label{final_pred}
% \vspace{-3mm}
\end{gather}
where $f_d$ is used to compute a new attentional hidden state $\tilde{s_j}$. Given the encoded input representations $\overleftrightarrow{H}=\{\overleftrightarrow{h_1}, ...,\overleftrightarrow{h_n}\}$ and the previous decoder GRU state $s_{j-1}$, the context vector at time step $j$ is computed as: $c_j= Attn(\overleftrightarrow{H}, s_{j-1})$.
% \begin{gather}
%     c_j= Attn(\overleftrightarrow{H}, s_{j-1})
%     \label{attn}
%     \vspace{-3mm}
% \end{gather}
 $Attn(\cdot,\cdot)$ computes a weight $\alpha_{jt}$ indicating the degree of relevance of an input text unit $x_t$ for predicting the target unit $y_j$ using a feed-forward network $f_{attn}$. 
%  Formally,
%  it includes a series of steps:
%     $a_{jt}=v_a^T$; $f_{attn}(s_{j-1},\overleftrightarrow{h_t})\\
%   \alpha_{jt}=\frac{exp(a_{jt})}{\sum_{k=1}^{n}{exp(a_{jk})}}$;
%     $c_j=\sum_{t}{\alpha_{jt}\overleftrightarrow{h_t}}$
%  \end{gather}
 Given a parallel corpus $D$, we train our model by minimizing the cross-entropy loss:
 $J=\sum_{(x,y)\in D}{-log p(y|x)}$.

\section{Adversarial Examples Generator (AEG) Architecture}
In this task of adversarial example generation, we have black-box access to the target model; the generator is not aware of the
target model architecture or parameters and is only capable of querying the target model with supplied inputs and obtaining the output predictions. To enable the model to have capabilities to generate word and character perturbations, we develop a hybrid encoder-decoder model, \textit{Adversarial Examples Generator (AEG)}, that operates at both word and character level to generate adversarial examples. Below, we explain the components of this model which have been improved to handle both word and character information from the text sequence.

%incorporating attention mechanism introduced by Bahdanau et al. \cite{}. Each text sequence can be represented in two ways: (a) a word sequence $[w_1, w_2,...,w_{L_w}$ and (b) a character sequence $[c_1,c_2,...,c_{L_c}]$. 

\subsection{Encoder}
\begin{figure}%{r}{0.7\linewidth}
\centering
\includegraphics[width=0.6\linewidth]{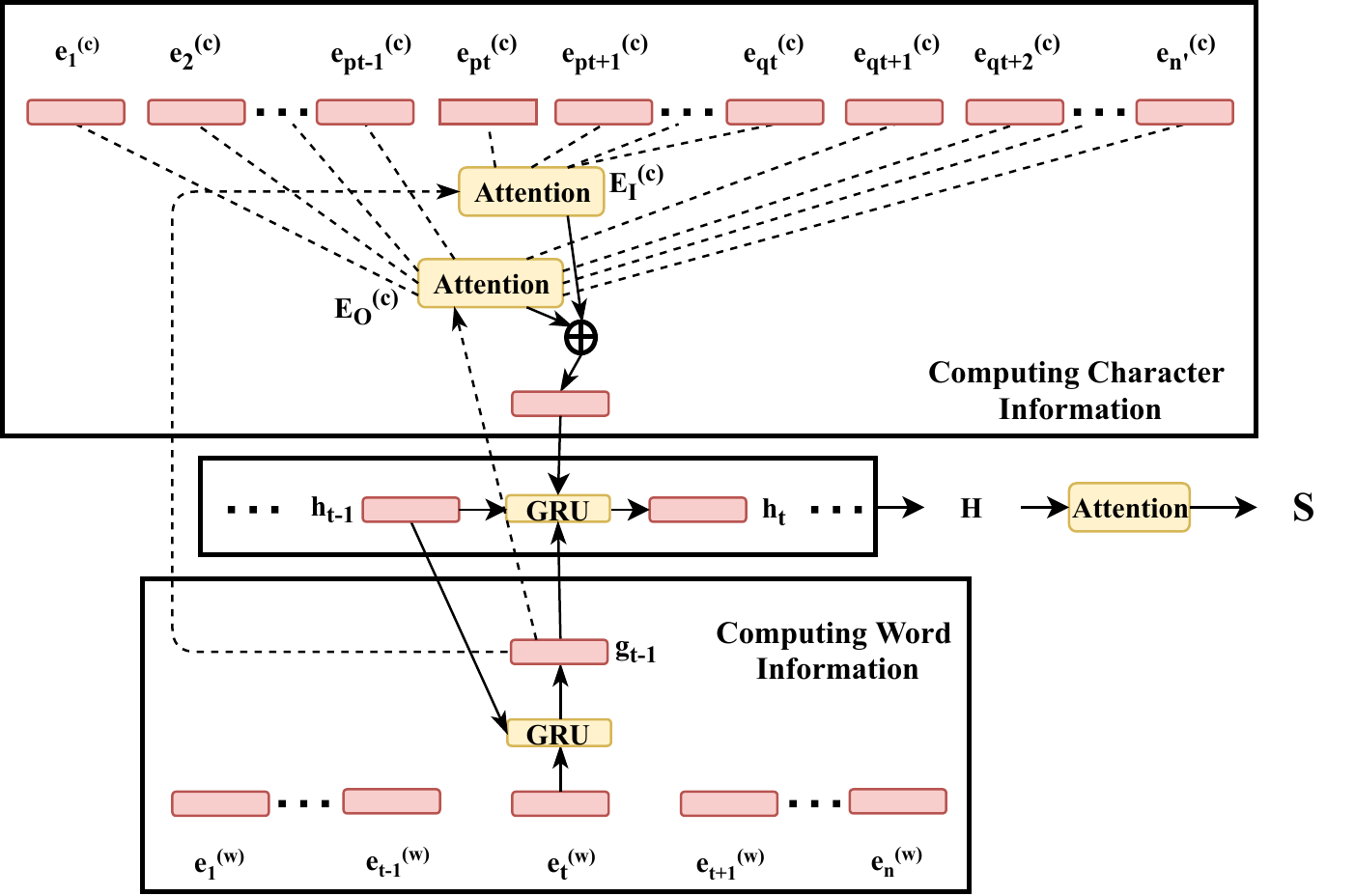}
\caption{Illustration of Encoder.}

\label{enco}
% \vspace{-4mm}
\end{figure}

The encoder maps the input text sequence into a sequence of representations using word and character-level information. Our encoder (Figure \ref{enco}) is a slight variant of Chen et al.\cite{chen2018combining}. This approach providing multiple levels of granularity can be useful in order to handle rare or noisy words in the text. Given character embeddings $E^{(c)}=[e_1^{(c)}, e_2^{(c)},...e_{n'}^{(c)}]$ and word embeddings  $E^{(w)}=[e_1^{(w)}, e_2^{(w)},...e_{n}^{(w)}]$ of the input, starting ($p_t$) and ending ($q_t$) character positions at time step $t$,  we define inside character embeddings as: $E_I^{(c)}=[e_{p_t}^{(c)},...., e_{q_t}^{(c)}]$ and outside embeddings as: $E_O^{(c)}=[e_{1}^{(c)},....,e_{p_t-1}^{(c)}; e_{q_t+1}^{(c)},...,e_{n'}^{(c)}]$. First, we obtain the character-enhanced word representation $\overleftrightarrow{h_t}$ by combining the word information from $E^{(w)}$ with the character context vectors. Character context vectors are obtained by attending over inside and outside character embeddings. Next, we compute a summary vector $S$ over the hidden states $\overleftrightarrow{h_t}$ using an attention layer expressed as $Attn(\overleftrightarrow{H})$. To generate adversarial examples, it is important to identify the most relevant text units that contribute towards the target model's prediction and then use this information during the decoding step to introduce perturbation on those units. Hence, the summary vector is optimized using target model predictions without back propagating through the entire encoder. This acts as a substitute network that learns to mimic the predictions of the target classifier.

\subsection{Decoder} 
\begin{figure}%{r}{0.7\linewidth}
\centering
\includegraphics[width=0.8\linewidth]{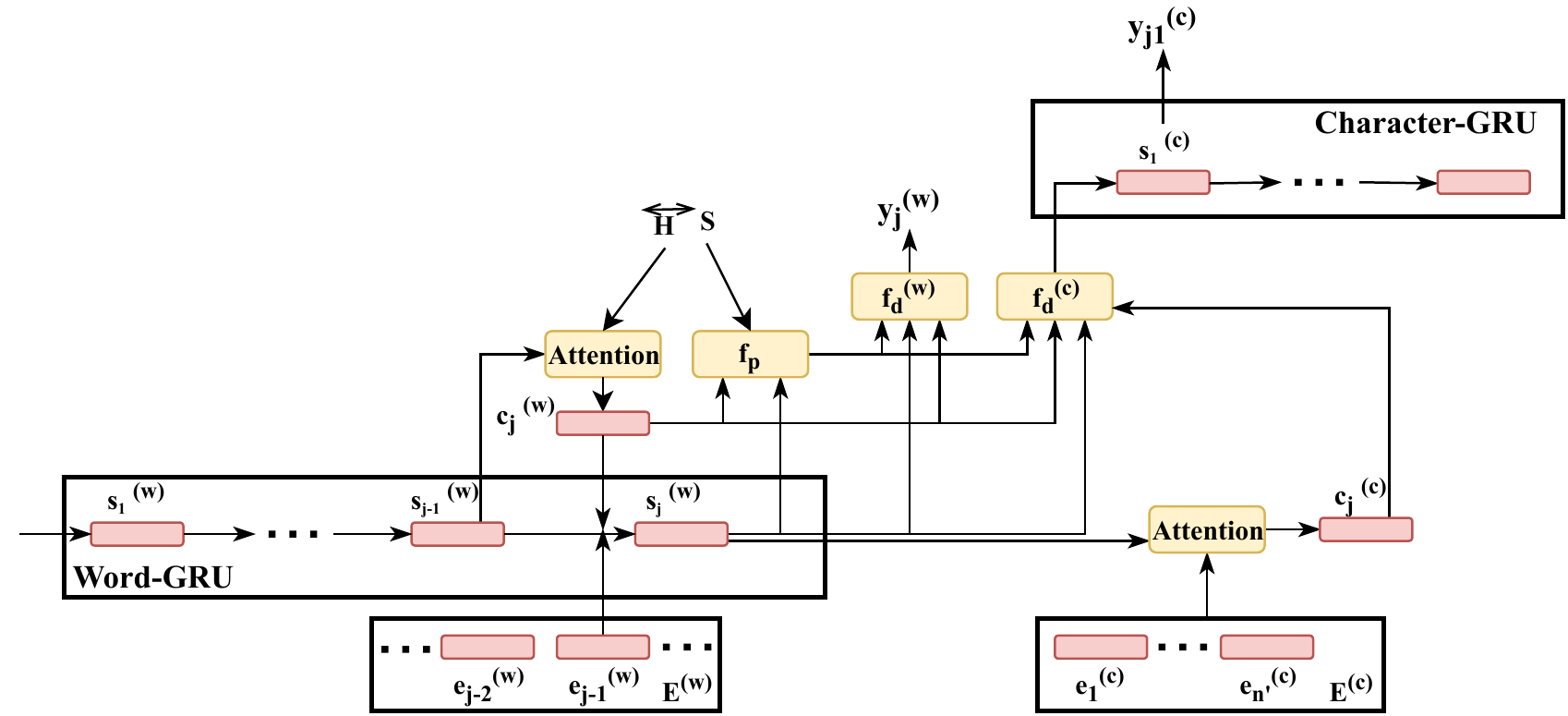}
\caption{Illustration of the word and character decoder.}

\label{deco}
% \vspace{-3mm}
\end{figure}

Our \textit{AEG} should be able to generate both character and word level perturbations as necessary. We achieve this by modifying the standard decoder \cite{bahdanau2014neural,luong2014addressing} to have two-level decoder GRUs: word-GRU and character-GRU (see Figure \ref{deco}). Such hybrid approaches have been studied to achieve open vocabulary NMT in some of the previous work like Wu et al. \cite{wu2016google} and Luong et al. \cite{luong2016achieving}. Given the challenge that all different word misspellings cannot fit in a fixed vocabulary, we leverage the power of both words and characters in our generation procedure. The word-GRU uses word context vector $c_j^{(w)}$  by attending over the encoder hidden states $\overleftrightarrow{h_t}$. Once the word context vector $c_j^{(w)}$ is computed, we introduce a \textit{perturbation vector} $v_{p}$ to impart information about the need for any word or character perturbations at this decoding step. We construct this vector using the word-GRU decoder state $s_j^{(w)}$, context vector $c_j^{(w)}$ and summary vector $S$ from the encoder as:
\begin{equation}
\setlength\abovedisplayskip{3pt}
v_{p}=f_{p}(s_j^{(w)},c_j^{(w)},S)
\label{pert}
\setlength\belowdisplayskip{3pt}
\end{equation}

We modify the the Equation \eqref{final_pred} as:
$\tilde{s}_j^{(w)}=f_{d}^{(w)}([c_j^{(w)};s_j^{(w)};v_{p}])$. The character-GRU will decide if the word is emitted with or without misspellings. We don't apply step-wise attention for character-GRU, instead we initialize it with the correct context. The ideal candidate representing the context must combine information about: (a) the word obtained from  $c_j^{(w)}, s_j^{(w)}$, (b) its character alignment with the input characters derived from character context vector $c_j^{(c)}$ with respect to the word-GRU's state and (c) perturbation embedded in $v_p$. This yields,
\begin{gather}
\setlength\abovedisplayskip{0pt}
    c_j^{(c)}=Attn(E^{(c)},s_j^{(w)})  \\
    \tilde{s}_j^{(c)}=f_{d}^{(c)}([c_j^{(w)};s_j^{(w)};v_{p};c_j^{(c)}]) 
\setlength\belowdisplayskip{0pt}
\end{gather}
Thus, $\tilde{s}_j^{(c)}$  is initialized to the character-GRU only for the first hidden state.  With this mechanism, both word and character level information can be used to introduce necessary perturbations.

% For example, take an simple perturbation of an input word ``happy'' is ``happyy''. Now, $\tilde{s}_j^{(w)}$ can be used to predict the word
\section{Training}
 \subsection{Supervised Pretraining with Teacher Forcing}
 The primary purpose of pretraining AEG is to enable our hybrid encoder-decoder to encode both character and word information from the input example and  produce both word and character-level transformations in the form of paraphrases or misspellings. Though the pretraining helps us mitigate the cold-start issue, it  does not guarantee that these perturbed texts will fool the target model. There are large number of valid perturbations that can be applied due to multiple ways of arranging text units to produce paraphrases or different misspellings. Thus, minimizing $J_{mle}$ is not sufficient to generate adversarial examples.
 
 \subsubsection{Dataset Collection}
 
 In this paper, we use paraphrase datasets like PARANMT-50M corpus\cite{wieting2017paranmt}, Quora Question Pair dataset \footnote{https://www.kaggle.com/c/quora-question-pairs/data} and Twitter URL paraphrasing corpus \cite{lan2017continuously}. These paraphrase datasets together contains text from various sources: Common Crawl, CzEng1.6, Europarl, News Commentary, Quora questions, and Twitter trending topic tweets. We do not use all the data for our pretraining. We randomly sample 5 million parallel texts and augment them using simple character-transformations (eg. random insertion, deletion or replacement) to words in the text. The number of words that undergo transformation is capped at 10\% of the total number of words in the text. We further include examples which contain only character-transformations without paraphrasing the original input. 
 
 \subsubsection{Training Objective}
  \textit{AEG} is pre-trained using teacher-forcing algorithm \cite{williams1989learning} on the dataset explained in Section \ref{datasets}. Consider an input text: ``movie was good'' that needs to be decoded into the following target perturbed text: ``film is gud''. The word ``gud'' might be out-of-vocabulary indicated by $<oov>$. Hence, we compute the loss incurred by word-GRU decoder, $J^{(w)}$, when predicting \{``film'', ``is'', ``$<oov>$''\} and loss incurred by character-GRU decoder, $J^{(c)}$, when predicting \{`f', `i',`l', `m', `\_'\},\{`i',`s','\_'\},\{`g', `u',`d',`\_'\}. Therefore, the training objective in Section \ref{decoder} is modified into:
  \begin{equation}
     J_{mle}=J^{(w)}+J^{(c)}
 \end{equation}

% \section{Learning Objective}
%  \subsection{Supervised Pre-training with Teacher Forcing}
%   \textit{AEG} is pre-trained using teacher-forcing algorithm \cite{williams1989learning} on the dataset explained in Section \ref{datasets}. Consider an input text: ``movie was good'' that needs to be decoded into the following target perturbed text: ``movie is gud''. The word ``gud'' might be out-of-vocabulary indicated by $<oov>$. Hence, we compute the loss incurred by word-GRU decoder, $J^{(w)}$, when predicting \{``film'', ``is'', ``$<oov>$''\} and loss incurred by character-GRU decoder, $J^{(c)}$, when predicting \{`f', `i',`l', `m', `\_'\},\{`i',`s','\_'\},\{`g', `u',`d',`\_'\}. Therefore, the training objective in Equation \eqref{obj} is modified into:
%  \begin{equation}
%      J_{mle}=J^{(w)}+J^{(c)}
%  \end{equation}

\subsection{Training with Reinforcement learning}
We fine-tune our model to fool a target classifier by learning a policy that maximizes a specific discrete metric formulated based on the constraints required to generate adversarial examples. In our work, we use the self-critical approach of Rennie et al. \cite{rennie2017self} as our policy gradient training algorithm.

\subsubsection{Self-critical sequence training (SCST)}
In SCST approach, the model learns to gather more rewards from its sampled sequences that bring higher rewards than its best greedy counterparts. First, we compute two sequences: (a) $y'$ sampled from the model's distribution $p(y'_j|y'_{<j},h)$ and (b) $\hat{y}$ obtained by greedily decoding ($argmax$ predictions) from the distribution $p(\hat{y}_j|\hat{y}_{<j},h)$ 
%for each training example at every timestep $j$ .
Next, rewards $r(y'_j),r(\hat{y}_j)$ are computed for both the sequences using a reward function $r(\cdot)$, explained in Section \ref{rewards}. We train the model by minimizing:
\begin{equation}
    J_{rl}=-\sum_j (r(y')-r(\hat{y}))log p(\hat{y}_j|\hat{y}_{<j},h)
\end{equation}
Here $r(\hat{y})$ can be viewed as the baseline reward. This approach, therefore, explores different sequences that produce higher reward compared to the current best policy.

 \subsubsection{Rewards}
 The reward $r(\hat{y})$ for the sequence generated is a weighted sum of different constraints required for generating adversarial examples.  Since our model operates at word and character levels, we therefore compute three rewards: adversarial reward, semantic similarity and lexical similarity reward. The reward should be high when: (a) the generated sequence causes the target model to produce a low classification prediction probability for its ground truth category, (b) semantic similarity is preserved and (c) the changes made to the original text are minimal. 
 
 \paragraph{Adversarial Reward} Given a target model $T$, it takes a text sequence $y$ and outputs prediction probabilities $P$ across various categories of the target model. Given an input sample $(x, l)$, we compute a perturbation using our AEG model and produce a sequence $y$. We compute the adversarial reward as $R_{A}=(1-P_l)$, where the ground truth $l$ is an index to the list of categories and $P_l$ is the probability that the perturbed generated sequence $y$ belongs to target ground truth $l$. Since we want the target classifier to make mistakes, we promote it by rewarding higher when the sequences produce low target probabilities.
 
 \paragraph{Semantic Similarity}
 Inspired by the work of Li et al. \cite{li2017paraphrase}, we train a deep matching model that can represent the degree of match between two texts. We use character based biLSTM models with attention \cite{lin2017structured} to handle word and character level perturbations. The matching model will help us compute the the semantic similarity $R_S$ between the text generated and the original input text.

 \paragraph{Lexical Similarity}
 Since our model functions at both character and word level, we compute the lexical similarity. The purpose of this reward is to keep the changes as minimal as possible to just fool the target classifier. Motivated by the recent work of Moon et al. \cite{moon2018multimodal}, we pretrain a deep neural network to compute approximate Levenshtein distance $R_{L}$ composed of character based bi-LSTM model. We replicate that model by generating a large number of text with perturbations in the form of insertions, deletions or replacements. We also include words which are prominent nicknames, abbreviations or inconsistent notations to have more lexical similarity. This is generally not possible using direct Levenshtein distance computation. Once trained, it can produce a purely lexical embedding of the text without semantic allusion. This can be used to compute the lexical similarity between the generated text $y$ and the original input text $x$ for our purpose.
 
Finally, we combine all these three rewards using:
 \begin{equation}
     r(y)=\gamma_A R_A+\gamma_S R_S+\gamma_L R_L
 \end{equation}
 where $\gamma_A, \gamma_S, \gamma_L$ are hyperparameters that can be modified depending upon the kind of textual generations expected from the model. The changes inflicted by different reward coefficients can be seen in Section \ref{result2}.

 \label{rewards}
 
 \subsection{Training Details}
 We trained our models on 4 GPUs. The parameters of our hybrid encoder-decoder were uniformly initialized to $[-0.1, 0.1]$. The optimization algorithm used is Adam \cite{kingma2014adam}. The encoder word embedding matrices were initialized with 300-dimensional Glove vectors \cite{pennington2014glove}. During reinforcement training, we used plain stochastic gradient descent with a learning rate of 0.01. Using a held-out validation set, the hyper-parameters for our experiments are set as follows: $\gamma_A=1, \gamma_S=0.5, \gamma_L=0.25$.

\section{Experiments}
In this section, we describe the evaluation setup used to measure the effectiveness of our model in generating adversarial examples. The success of our model lies in its ability to fool the target classifier. We pretrain our models with dataset that generates a number of character and word perturbations. We elaborate on the experimental setup and the results below.

\begin{table*}[htbp]
  \centering
  \begin{tabular}{l|l|l|l}
    \toprule
    \textbf{Datasets} & \textbf{Details} & \textbf{Model} & \textbf{Accuracy} \\
    \midrule
    IMDB Review &  Classes: 2; \#Train: 25k; & CNN-Word \cite{kim2014convolutional} & 89.95\%\\ 
    AG's News  & Classes: 4; \#Train: 120k; & CNN-Char \cite{zhang2015character} & 89.11\%  \\
    \bottomrule
  \end{tabular}
  \caption{Summary of data and models used in our experiments.}
  \label{table:datasets}
%   \vspace{-7mm}
\end{table*} 
\subsection{Setup}
We conduct experiments on different datasets to verify if the accuracy of the deep learning models decrease when fed with the adversarial examples generated by our model. We use benchmark sentiment classification and news categorization datasets and the details are as follows: 
\begin{itemize}
\item Sentiment classification: We trained a word-based convolutional model (CNN-Word) \cite{kim2014convolutional} on IMDB sentiment dataset \footnote{http://ai.stanford.edu/~amaas/data/sentiment/}. The dataset contains 50k movie reviews in total which are labeled as positive or negative. The trained model achieves a test accuracy of 89.95\% which is relatively close to the state-of-the-art results on this dataset. 

\item News categorization: We perform our experiments on AG's news corpus \footnote{https://github.com/mhjabreel/CharCNN/tree/master/data/ag\_news\_csv} with a character-based convolutional model (CNN-Char) \cite{zhang2015character}. The news corpus contains titles and descriptions of various news articles along with their respective categories. There are four categories: World, Sports, Business and Sci/Tech. The trained CNN-Char model achieves a test accuracy of 89.11\%.
\end{itemize}
Table \ref{table:datasets} summarizes the data and models used in our experiments. We compare our proposed model with the following black-box non-targeted attacks:
\begin{itemize}
    \item \textbf{Random}: We randomly select a word in the text and introduce some perturbation to that word in the form of a character replacement or synonymous word replacement. No specific strategy to identify importance of words.
    \item \textbf{NMT-BT}: We generate paraphrases of the sentences of the text using a back-translation approach \cite{iyyer2018adversarial}. We used pretrained English$\leftrightarrow$German translation models to obtain back-translations of input examples.
    \item \textbf{DeepWordBug} \cite{gao2018black}: A scoring function is used to determine the important tokens to change. The tokens are then modified to evade a target model. 
    \item \textbf{No-RL}: We use our pretrained model without the reinforcement learning objective.

\end{itemize}
The performance of these methods are measured by the percentage fall in accuracy of these models on the generated adversarial texts. Higher the percentage dip in the accuracy of the target classifier, more effective is our model.

\subsection{Quantitative Analysis}
We analyze the effectiveness of our approach by comparing the results from using two different baselines against character and word-based models trained on different datasets. Table \ref{results}  demonstrates the capability of our model. Without the reinforcement learning objective, the No-RL model performs better than the back-translation approach(NMT-BT). The improvement can be attributed to the word and character perturbations introduced by our hybrid encoder-decoder model as opposed to only paraphrases in the former model. Our complete \textit{AEG} model outperforms all the other models with significant drop in accuracy. For the CNN-Word, DeepWordBug decreases the accuracy from 89.95\% to 28.13\% while \textit{AEG} model further reduces it to 18.5\%.
% \begin{table}[ht!]
%   \centering
%   \begin{tabular}{c|c|c}
%     \toprule
%     \textbf{Models} & \textbf{IMDB} & \textbf{AG's News } \\
%      & (CNN-Word) & (CNN-Char) \\
%     \midrule
%     Random  &  2.46\%  & 9.64\%  \\ 
%     NMT-BT  & 25.38\% &  22.45\% \\
%     DeepWordBug & 68.73\% &  65.80\% \\
%     No-RL (Ours) & 38.05\% &  33.58\% \\
%     \textbf{AEG (Ours)}  & \textbf{79.43\%} & \textbf{72.16\%}  \\
%     \bottomrule
%   \end{tabular}
%   \caption{Performance of our AEG model on IMDB and AG's News dataset using word and character based CNN models respectively. Results indicate the percentage dip in the accuracy by using the corresponding attacking model over the original accuracy.}
%   \label{results}
% \end{table}

\begin{table}[!htb]
    
    \subfloat{
      \centering
        \begin{tabular}{c|c|c}
    \toprule
    \textbf{Models} & \textbf{IMDB} & \textbf{AG's News } \\
     & (CNN-Word) & (CNN-Char) \\
    \midrule
    Random  &  2.46\%  & 9.64\%  \\ 
    NMT-BT  & 25.38\% &  22.45\% \\
    DeepWordBug & 68.73\% &  65.80\% \\
    No-RL (Ours) & 38.05\% &  33.58\% \\
    \textbf{AEG (Ours)}  & \textbf{79.43\%} & \textbf{72.16\%}  \\
    \bottomrule
  \end{tabular}
  }
    \subfloat{
      \centering
    %   \caption{}
        \begin{tabular}{|c|c|c|}
            \hline Model Variants & IMDB & News Corpus  \\ \hline
            Char-dec & 73.5 & 68.64\%  \\
            No pert & 71.45\% & 65.91\%  \\
            \hline
        \end{tabular}
    }
    
\caption{\textbf{Left:} Performance of our AEG model on IMDB and AG's News dataset using word and character based CNN models respectively. Results indicate the percentage dip in the accuracy by using the corresponding attacking model over the original accuracy. \textbf{Right:} Performance of different variants of our model.}
\label{results}
% \vspace{-7mm}
\end{table}

It is important to note that our model is able to expose the weaknesses of the target model irrespective of the nature of the model (either word or character level). It is interesting that even simple lexical substitutions and paraphrases can break such models on both datasets we tested. Across different models, the character-based models are less susceptible to adversarial attacks compared to word-based models as they are able to handle misspellings and provide better generalizations.

\subsection{Human Evaluation}
We also evaluated our model based on human judgments. We conducted an experiment where the workers were presented with randomly sampled 100 adversarial examples generated by our model which were successful in fooling the target classifier. The examples were shuffled to mitigate ordering bias, and every example was annotated by three workers. The workers were asked to label the sentiment of the sampled adversarial example. For every adversarial example shown, we also showed the original text and asked them to rate their similarity on a scale from $0$ (Very Different) to $3$ (Very Similar). We found that the perturbations produced by our model do not affect the human judgments significantly as $94.6\%$ of the human annotations matched with the ground-truth label of the original text. The average similarity rating of $1.916$ also indicated that the generated adversarial sequences are semantics-preserving.

%  \begin{table}[h!]
% \small
% \begin{center}
% \begin{tabular}{|c|c|c|}
% \hline Model Variants & IMDB & News Corpus  \\ \hline
% Char-dec & 73.5 & 68.64\%  \\
% No pert & 71.45\% & 65.91\%  \\
% \hline
% \end{tabular}
% \end{center}
% \caption{Performance of different variants of our model.}
% \label{ablation}
% \normalsize
% \end{table}
 \subsection{Ablation Studies}
\begin{figure*}%{r}{0.7\linewidth}
\centering
\includegraphics[width=\linewidth]{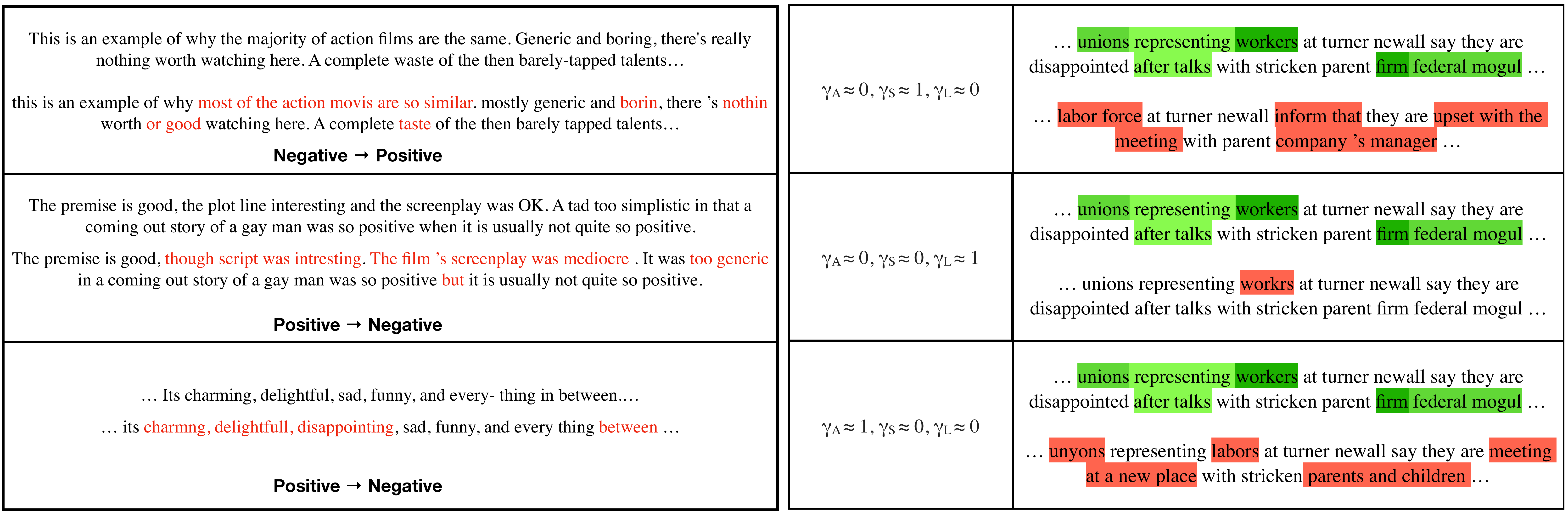}
\caption{\textbf{Left}: Examples from IMDB reviews dataset, where the model introduces misspellings or paraphrases that are sufficient to fool the target classifier. \textbf{Right}: Effect of coefficients of the reward function. The first line is the text from the AG's news corpus. The second line is the generated by the model given specific constraints on the reward coefficients. The examples do not necessarily lead to misclassification. The text in green are attention scores indicating relevance of classification. The text in red are the perturbations introduced by our model.}

\label{scores}
% \vspace{-3mm}
\end{figure*}

 In this section, we make different modifications to our encoder and decoder to weigh the importance of these techniques: (a) No perturbation vector (No Pert) and finally (b) a simple character based decoder (Char-dec) but involves perturbation vector. Table \ref{results} shows that the absence of hybrid decoder leads to a significant drop in the  performance of our model. The main reason we believe is that hybrid decoder is able to make targeted attacks on specific words which otherwise is lost while generating text using a  pure-character based decoder. In the second case case, the most important words associated with the prediction of the target model are identified by the summary vector. When the perturbation vector is used, it carries forward this knowledge and decides if a perturbation should be performed at this step or not. This can be verified even in Figure \ref{scores}, where the regions of high attention get perturbed in the text generated. 

% The main reason for choosing a word and character encoder is also justified by the need for extracting information at both the levels simultaneously. This encoder offers to separate itself from other CNN+RNN models or other word-character model because the word boundaries are clearly delineated when we process the characters through our encoders. It fuses the character information together with the word embeddings at every time step. There are other word and character models where the fusion happens at a later stage. We find this kind of model more  promising from the results in Tables \ref{results} and \ref{ablation}.
\subsection{Qualitative Analysis}
\label{result2}
We qualitatively analyze the results by visualizing the attention scores and the perturbations introduces by our model. We further evaluate the importance of hyperparameters $\gamma_{(.)}$ in the reward function. We set only one of the hyperparameters closer to 1 and set the remaining closer to zero to see how it affects the text generation. The results can be seen in Figure \ref{scores}. Based on a subjective qualitative evaluation, we make the following observations:

\begin{itemize}
    \item Promisingly, it identifies the most important words that contribute to particular categorization. The model introduces misspellings or word replacements without significant change in semantics of the text.
    \item When the coefficient associated only with adversarial reward goes to 1, it begins to slowly deviate though not completely. This is motivated by the initial pretraining step on paraphrases and perturbations.
\end{itemize}

\section{Conclusion}
In this work, we have introduced a $AEG$, a model capable of generating adversarial text examples to fool the black-box text classification models. Since we do not have access to gradients or parameters of the target model, we modelled our problem using a reinforcement learning based approach. In order to effectively baseline the REINFORCE algorithm for policy-gradients, we implemented a self-critical approach that normalizes the rewards obtained by sampled sentences with the rewards obtained by the model under test-time inference algorithm. By generating adversarial examples for target word and character-based models trained on IMDB reviews and AG's news dataset, we find that our model is capable of generating semantics-preserving perturbations that leads to steep decrease in accuracy of those target models.
We conducted ablation studies to find the importance of individual components of our system. Extremely low values of the certain reward coefficient constricts the quantitative performance of the model can also lead to semantic divergence. Therefore, the choice of a particular value for this model should be motivated by the demands of the context in which it is applied. One of the main challenges of such approaches lies in the ability to produce more synthetic data to train the generator model in the distribution of the target model's training data. This can significantly improve the performance of our model. We hope that our method motivates a more nuanced exploration into generating adversarial examples and adversarial training for building robust classification models.

\bibliographystyle{splncs04}
\bibliography{llncs}





% LNCS Homepage, \url{http://www.springer.com/lncs}. Last accessed 4
% \end{thebibliography}
\end{document}